# Effects of Stop Words Elimination for Arabic Information Retrieval: A Comparative Study

**Ibrahim Abu El-Khair**
Assistant Professor
Dept. of Library and Information Science
Faculty of Arts, El-Minia University-Egypt
iabuelkhair@gmail.com

***Abstract:*** *The effectiveness of three stop words lists for Arabic Information Retrieval---General Stoplist, Corpus-Based Stoplist, Combined Stoplist ---were investigated in this study. Three popular weighting schemes were examined: the inverse document frequency weight, probabilistic weighting, and statistical language modelling. The Idea is to combine the statistical approaches with linguistic approaches to reach an optimal performance, and compare their effect on retrieval. The LDC (Linguistic Data Consortium) Arabic Newswire data set was used with the Lemur Toolkit. The Best Match weighting scheme used in the Okapi retrieval system had the best overall performance of the three weighting algorithms used in the study, stoplists improved retrieval effectiveness especially when used with the BM25 weight. The overall performance of a general stoplist was better than the other two lists.*

## 1. Introduction

Although most of the research in the field of information retrieval has focused on the English language, recently there has been a considerable amount of work and effort to develop information retrieval systems for languages other than English. Research and experimentation in the field of information retrieval in the Arabic language is relatively new and limited compared to the research that has been done in English, which has been dominant in the field of information retrieval for a long while. This is despite the fact that the Arabic Language is one of the five languages of the United Nations, the mother tongue of over 256 million people. In addition, because it is the language of the Qur'an, it is also the second language for many Moslems and Moslem countries around the world [4].

This study attempts to compare the use and effect of stop words for Arabic information retrieval. Using the Lemur Toolkit, a language modeling and information retrieval package (see Methodology for more details), multiple weighting schemes, and three stopword lists are implemented in order to determine the effect of stop words elimination on an Arabic information retrieval system.

The weighting schemes to be used are the TF*IDF weight, the best match weight (BM25), and the statistical language modelling (KL). Three stop words lists will be created, a general list, a corpus-based list, and a combined list. Although Stemming is an important factor when dealing with Arabic information retrieval, it was not implemented in this study in order to isolate the effect of stop words from any other factor.

## 2. Related Studies

Stopwords are very common words that appear in the text that carry little meaning; they serve only a syntactic function but do not indicate subject matter. These stopwords have two different impacts on the information retrieval process. They can affect the retrieval effectiveness because they have a very high frequency and tend to diminish the impact of frequency differences among less common words, affecting the weighting process. The removal of the stopwords also changes the document length and subsequently affects the weighting process. They can also affect efficiency due to their nature and the fact that they carry no meaning, which may result in a large amount of unproductive processing [9]. The removal of the stopwords can increase the efficiency of the indexing process as 30 to 50% of the tokens in a large text collection can represent stopwords [18].

Identifying a stopwords list or a *stoplist* that contain such words in order to eliminate them from text processing is essential to an information retrieval system. Stoplists can be divided into two categories; domain independent stoplists and domain dependent stoplists. They can be created using syntactic classes or using corpus statistics, which is a more domain dependent approach, used for well-defined fields. They can also be created using a combination of the syntactic classes and corpus statistics to obtain the benefits of both approaches.

Fox [6] was the first to create an English stoplist to be used for general text based on word usage in English. He generated a stoplist that consisted of 421 words which was used later with in the Okapi retrieval system. Fox's method in creating the list is the most frequently used method; it is a domain independent approach. The problem with this method is that there are several arbitrary decisions to be taken during the creation of the list, such as the cut-off point. The elimination of some words and addition of others is based on personal judgment, which requires a certain expertise with the language in hand.

There is no general standard stoplist to use in an IR experiment for the Arabic language. The stoplist used in the Lemur Toolkit is the one created by Khoja [8] when she was creating her Arabic stemmer and is relatively short (168 words). This list was used by Larkey and Connell [11] and Larkey, Ballesteros and Connell [10]. Chen and Gey [5] used a list they created by translating an English list and augmenting it with high frequency words from the corpus creating a rather large list, 1,131 words. They do not discuss the effect of the list.

Savoy and Rasolofo's stoplist [1] [17] is a domain dependent list which has three problems. First they use some words preceded by the letter waw "و" which means "and" in 17 words including 11 duplicates. This letter comes in its separate format in a large portion of words in the Arabic language and could precede all the words in the language with no exceptions. A more efficient way to do this is to remove it using a good stemming algorithm. Second, they remove several other single letters with the waw namely the hamza "ء", alef "ا, أ", ba' "ب", heh "ه". Due to the way the Arabic language is written, these letters can come separately but they are still a part of the word and removing them changes the word meaning or leave it meaningless, e.g. the word "كتّاب" which could mean book, writers, or a place for learning has the letter ba' as a single separate letter and when it is removed the word is meaningless. The third problem is that some of the words used in it are not stopwords even though they appeared frequently in the analysis of the corpus statistics, for example, "الولايات" States, "المتحدة" United, "القاهرة" Cairo, etc. In addition, it is a more domain dependent list so it may not be suitable for other collections.

[1] The stoplists for all the languages are available at http://www.unine.ch/info/clef

## 3. Methodology

This study explores the use of stop words and their effect on Arabic information retrieval. It compares the use of three term weighting schemes, and three stoplists. These techniques are examined using a large corpus that was not available before the introduction of Arabic Cross-Language Retrieval at TREC 2001. The evaluation used the Lemur Toolkit with Arabic language capability.

The study evaluates these techniques using the standard recall and precision measures as the basis for comparison. It answers the following question:

- What is the effect of the stoplists on retrieval, i.e. how sensitive is retrieval to the use of stopwords; and which one of the lists, the general, the corpus based, or the combined list is superior to the other?

First, performance of term weighting schemes without elimination of stopwords was compared, and then combinations of weighting schemes, and stoplists were run. Using statistical analysis, the effectiveness of all techniques was evaluated to determine which combination achieves the optimal performance for Arabic language retrieval.

### *3.1. Data Set*

This research used one Arabic test corpus, created in the Linguistic Data Consortium in Philadelphia, also used in the recent TREC experiments. The Arabic Newswire A corpus was created by David Graff and Kevin Walker at the Linguistic Data Consortium [13]. It is composed of articles from the Agence France Presse (AFP) Arabic Newswire. The source material was tagged using TIPSTER style SGML and was transcoded to Unicode (UTF-8). The corpus includes articles from 13 May 1994 to 20 December 2000. The data is in 2,337 compressed Arabic text data files. There are 209 Mbytes of compressed data (869 Mbytes uncompressed) with 383,872 documents containing 76 Million tokens over approximately 666,094 unique words.

### *3.2. Query Sets and Relevance Judgments*

The query set associated with the LDC corpus was created for TREC 2001 and 2002 [12, 20, 21]. It consists of 75 queries, developed at the LDC by native Arabic speakers and translated to English and French. The relevance judgements for these queries were obtained using assessment pools from different runs at TREC 2001 and 2002, and using the top 70 documents from each run with an average size of 910 documents for each pool. For TREC 2001, the average number of relevant documents over the 25 queries was 164.9 with five topics having more than 300 relevant documents and another five with fewer than 25 relevant documents [22]. For TREC 2002, the average number of relevant documents over the 50 queries was 118.2 with eight topics having more than 300 relevant documents and 16 topics with less than 25 relevant documents.

### *3.3. Retrieval Engine*

The Lemur Toolkit for Language Modelling and Information Retrieval was used. The results of the experiments were mapped against the relevance judgments that are available for the data set. Standard recall and precision measures were calculated using the *ireval.pl* routine in the Toolkit. The evaluation was based on the use of eleven levels of recall creating the recall/precision matrix.

The Lemur toolkit was chosen for several reasons. It supports the construction of basic text retrieval systems using language modelling methods, as well as traditional methods such as those based on the vector space model and Okapi. It is available on the Web as open source software written in C and C++, and runs on both UNIX and Windows (NT). It was developed by collaboration between the Computer Science Department at the University of Massachusetts and the School of Computer Science at Carnegie Mellon University.

Arabic language capability was recently added to the Toolkit by Leah Larkey. This addition has solved the problem of the availability of an Arabic retrieval system for research and experimentation. The Toolkit uses Windows CodePage 1256 encoding (CP1256).

The Toolkit comes equipped with *TextQueryRetMethods* that implement a basic TF*IDF vector space model, Okapi, and a language modeling method using the Kullback-Leibler similarity measure between document and query language models. Initially these algorithms were used without stemming or stoplists. Each of these algorithms requires several parameters to be set in order to function properly. The parameters that were used were the default parameters set in Lemur. Several unofficial runs were conducted in an attempt to tune these parameters to the collection in hand but this did not improve the results over the defaults in Lemur.

These parameters are as follows: TF*IDF parameters: K1 = 1, B = 0.3; BM25 Parameters: K1 = 1.2, B = 0.75, K3 = 7; KL-Divergence Model parameter: The KL model needs only one parameter that is used for the smoothing algorithm applied in it. This study uses the simple KL model with the Dirichlet Prior smoothing algorithm and the Dirichlet Prior parameter was set to 2000 as a typical value for that parameter since currently there is no good way of estimating it [23].

### *3.4. Stoplists*

A general stoplist was created, based on the Arabic language structure and characteristics without any additions. All possible words or articles that may be considered a stopword were collated from the different syntactic classes in Arabic in a systematic way to ensure the completeness of the list.
The word categories [1, 2] that were used are:

- Adverbs.
- Conditional Pronouns.
- Interrogative Pronouns.
- Prepositions.
- Pronouns.
- Referral Names/Determiners.
- Relative Pronouns
- Transformers (verbs, letters).
- Verbal Pronouns.
- Other.

Choosing a word from any of these categories was based on a personal judgment. Not all the words under these categories were used, as some of them were not considered stopwords.

The resulting list consisted of ***1,377 words***. The list was checked against Khoja [8] and Alshehri’s [3] lists, and two Standard English lists, the Okapi and SMART lists. The reason for having a large number of stopwords is due to the characteristics of the Arabic language. First, in Arabic several letters can be used as prefixes and may change the meaning of the word. These letters are ("ل" "ك" "ف" "ب" "أ"), and they were used on some of the words, not all of them, that they could be used with. Second, a considerable number of the original words from these categories could be joined together and used as suffixes or prefixes, especially the pronouns. Finally, the conjunction letter WAW meaning “and” could be used in the same way but was not because it could be used for all Arabic words with no exception and it would not be realistic to use it.

In order to test the effect of a corpus-based stoplist, a second list was created. A cut-off point determining a certain number of words at which the list will stop, was decided based on the corpus statistics. Words occurring more than 25,000 times were used to create this list. Preliminary examination of the corpus word-frequency statistics showed that this is a reasonable number to use as the cut-off point, even though it may appear to be an arbitrary decision.

Under this condition, 359 words occurred with a frequency of more than 25,000. Then doing a manual check to remove any content bearing word, which may not be considered a conventional stopword, from this list. The removal of these words is another arbitrary decision based on personal judgment, and the reason for doing this is that there are no clear rules on stopwords list creation. The resulting list contained ***235 words***.
A third stoplist created using both the general and corpus-based stoplist. Combining both lists resulted in a list of ***1,529 words***. Of these 83 words were in common between the two lists.

### 3.5. Experimental Setup

The data and the query set for the experiments were processed as follows.

- The 383,872 files in the data set were converted from *UTF-8* format to Windows Code Page 1256 encoding (*CP1256*) for Lemur compatibility.
- The queries were converted from the *ASMO 708* encoding to *CP1256* encoding.
- Title and description for each of the 75 queries were extracted from the original query set.
- Several fatal spelling errors in the queries were corrected.
- The table Normalization function in the Light-10 stemmer in Lemur was implemented for all the runs regardless of the techniques that were used.
    - The letters (أ، إ، آ) were replaced with (ا).
    - The final (ى) was replaced with (ي).
    - The Final (ة) was replaced with (ه).
- In order to avoid confusion each technique was given a code to represent it throughout the experiments:
    - Term Frequency Weighting Scheme: *TFIDF*.
    - Okapi Weighting Scheme: *BM25*.
    - Language Modeling (Kullback-Leibler Divergence Model): *KL*.
    - General Stoplist: *GS*.
    - Corpus-Based Stoplist: *CBS*.
    - Combined Stoplist: *CS*.

Combinations of the techniques were coded starting with the weighting scheme, and then the stoplist. For instance, *BM25_CBS* represents a combination of the Okapi Weighting Scheme, and the Corpus-Based Stoplist.

### 3.6. Evaluation

The performance of each technique was evaluated using the standard measures of *Recall* and *Precision,* [9, 15]. A total of 27 runs were carried out where each run represents one or a combination of more than one of these techniques. The raw output results obtained from the *RetEval* application in Lemur were processed with the *ireval.pl* script to give recall and precision. The script does a TREC-style evaluation and the output includes:

- Total number of relevant documents.
- Total number of relevant documents retrieved.
- Average non-interpolated precision.
- Interpolated precision over the eleven levels of recall.
- Non-interpolated precision at document cut-off levels.
- Breakeven point (exact) precision.

### 3.7. Data Analysis

Retrieval results were analyzed by calculating the differences between the Recall and Precision scores, and plotting them in the R-P graph. The *Friedman Two-Way ANOVA test* and the *Wilcoxon Matched-Paired Signed-Rank test* were used for judging whether measured differences between different methods can be considered statistically significant or not.

Hull [7] has examined the validity of different statistical techniques that are used in comparing retrieval experimental results. He states that there are two non-parametric alternatives to the t-test that make no assumptions about the shapes of the distributions of the two variables, the *Wilcoxon Matched-Paired Signed-Rank test* and the sign test.

The sign test looks only at the sign of the difference, ignoring its magnitude. If one method performs better than the other far more frequently than would be expected on average, then this is strong evidence that it is superior. The Wilcoxon test replaces each difference with the rank of its absolute value. These ranks are then multiplied by the sign of the difference, and the sum of the ranks for each group is compared to its expected value under the assumption that the two groups are equal.

The reason for choosing the Wilcoxon signed rank test over the sign test is that it is a more powerful and indicative test as it considers the relative magnitude in addition to the direction of the differences considered [19]. It also assumes that as differences between pairs increase, significance also increases [16].

Hull also indicates that when comparing more than two retrieval methods, the *Friedman Two-Way ANOVA* is appropriate. It is a non-parametric equivalent to the One-Way ANOVA that does not require any assumptions. It is used to compare observations repeated on the same subjects, which is the case in hand. It uses the ranks of the data rather than their raw values to calculate the statistic [19].

In this study the *Friedman Two-Way ANOVA test* was used to indicate if there is a significant difference on multiple techniques, then it was followed by the *Wilcoxon Matched-Paired Signed-Rank test* which was used to test pair-wise differences.

# 4. Results and Data Analysis

This study compared alternative stop word list and their effect on retrieval effectiveness. Six different techniques with a total of 12 different combinations were examined. Results are compared using the Wilcoxon test (see Table 2), and recall and precision curves (figures 1-4).

The Friedman Two-way ANOVA test was used to determine if the differences are statistically significant (see table 1), the test statistic $\chi^2$ = 70.471 and the P-value* = 0.000 which indicates that the differences between techniques as a whole are statistically significant. This was not surprising considering the wide variety of techniques used but the test does not depict individual differences between two techniques.

**Table 1:** The Friedman Test for All Techniques

| Technique | Mean Precision | Mean Rank |
|---|---|---|
| ***BM25*** | ***0.2169*** | ***4.91*** |
| KL_CBS | 0.2217 | 4.95 |
| KL_CS | 0.2243 | 5.76 |
| ***KL*** | ***0.2264*** | ***6.04*** |
| TFIDF_CBS | 0.2260 | 6.23 |
| KL_GS | 0.2248 | 6.29 |
| ***TFIDF*** | ***0.2260*** | ***6.42*** |
| TFIDF_GS | 0.2283 | 7.09 |
| BM25_CBS | 0.2433 | 7.11 |
| TFIDF_CS | 0.2285 | 7.3 |
| BM25_CS | 0.2474 | 7.73 |
| BM25_GS | 0.2496 | 8.45 |

In order to explore these differences, the techniques are grouped according to weighting scheme, and stoplists and combinations of them. For each group of retrieval techniques the significance testing starts with the Friedman Two-way ANOVA test to observe the differences between all techniques used are statistically significant. Then, a post-hoc test using the Wilcoxon Matched-Paired Signed-Rank test was performed to determine the difference between paired techniques. Due to the number of techniques used and the number of different combinations of them, it was impractical to do a pair-wise comparison on all of them. To set a baseline for comparison, the raw term weighting techniques were run without any additional techniques and the best of the three was used as the baseline.

## *4.1. Term Weighting*

The results of the term weighting approach show that the three algorithms performed relatively well considering the difficulties of the Arabic language and the fact that no linguistic adaptation for it was implemented during retrieval.

Although the BM25 and the KL model are known to perform well compared to the TFIDF weight, in the current study, the overall performance of TFIDF weight was better than the performance of both the BM25 and KL model (see figure 1). The good performance of TFIDF is due to the way the term frequency portion of the weight is calculated in the Lemur Toolkit, using the TF function from the BM25 scheme, which improves its performance significantly.

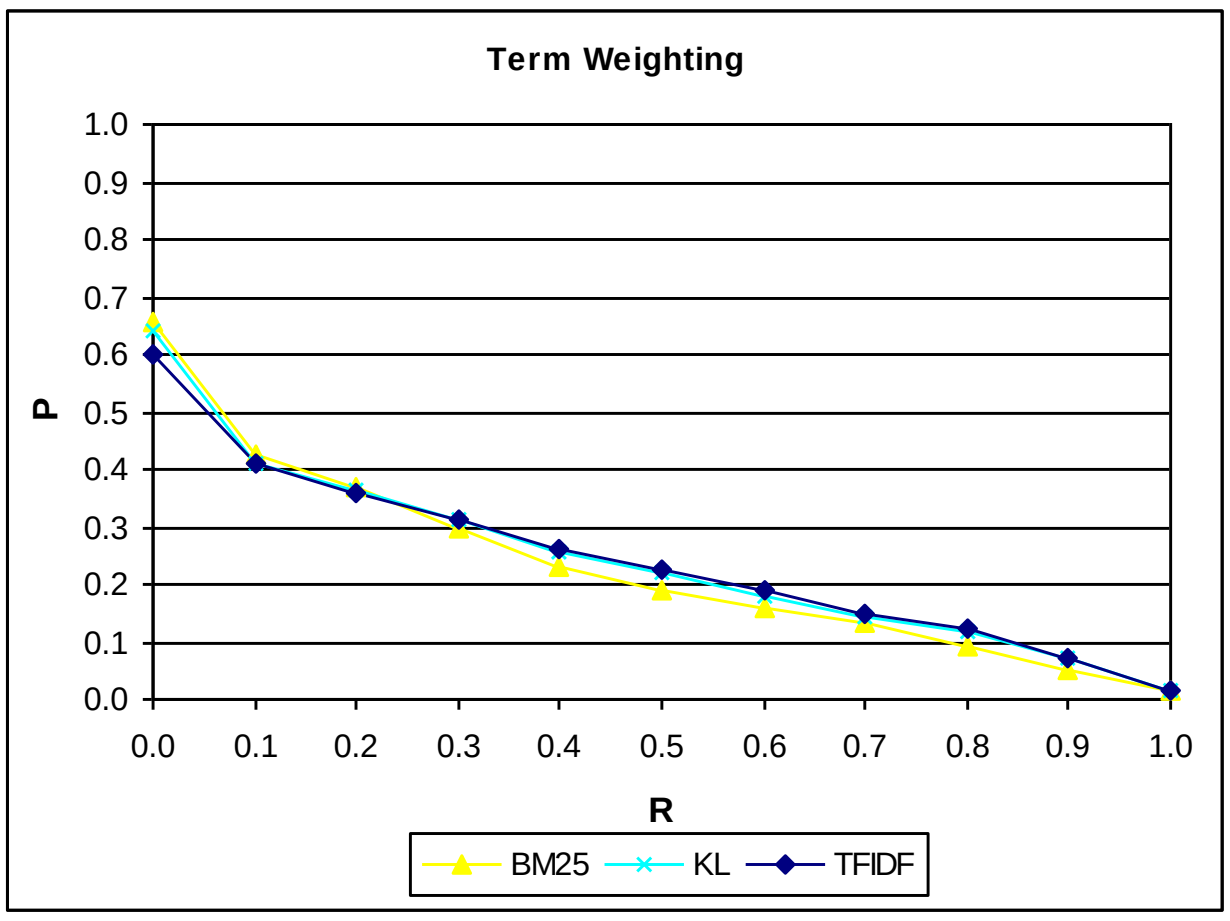

**Figure 1:** Recall-Precision Curves for Term Weighting

In the Friedman test, $\chi^2$ = 5.946; P-value = .051 and the P-value indicates that the differences between these three techniques are not statistically significant. The Wilcoxon test was used to determine if the pair-wise differences are statistically significant.
Because the TFIDF had the best overall performance, it was used as the baseline for comparisons. The Wilcoxon test confirms the result of the Friedman test results, the differences between TFIDF and BM25, and TFIDF and KL were not statistically significant.

## *4.2. Term Weighting and Stoplists*

This subsection presents the results obtained by using the three stoplists that were created for this study. The stoplists were created assuming that they would improve the retrieval efficiency when used with other techniques. The results illustrate how sensitive retrieval is to the use of stopwords. The stopwords will essentially affect the term weights used as they have a significant effect on the term frequency.

### *4.2.1. General Stoplist*

After creating the list, it was initially used in combination with term weighting. The results obtained when using the general stoplist are presented in figure 2. The test statistic for the Friedman test $\chi^2$ = is 14.517 and the P-value is .002. This indicates that the differences between these runs and the baseline precision are significant.

* The P-Value for both Wilcoxon and Friedman tests was set on the .05 level.

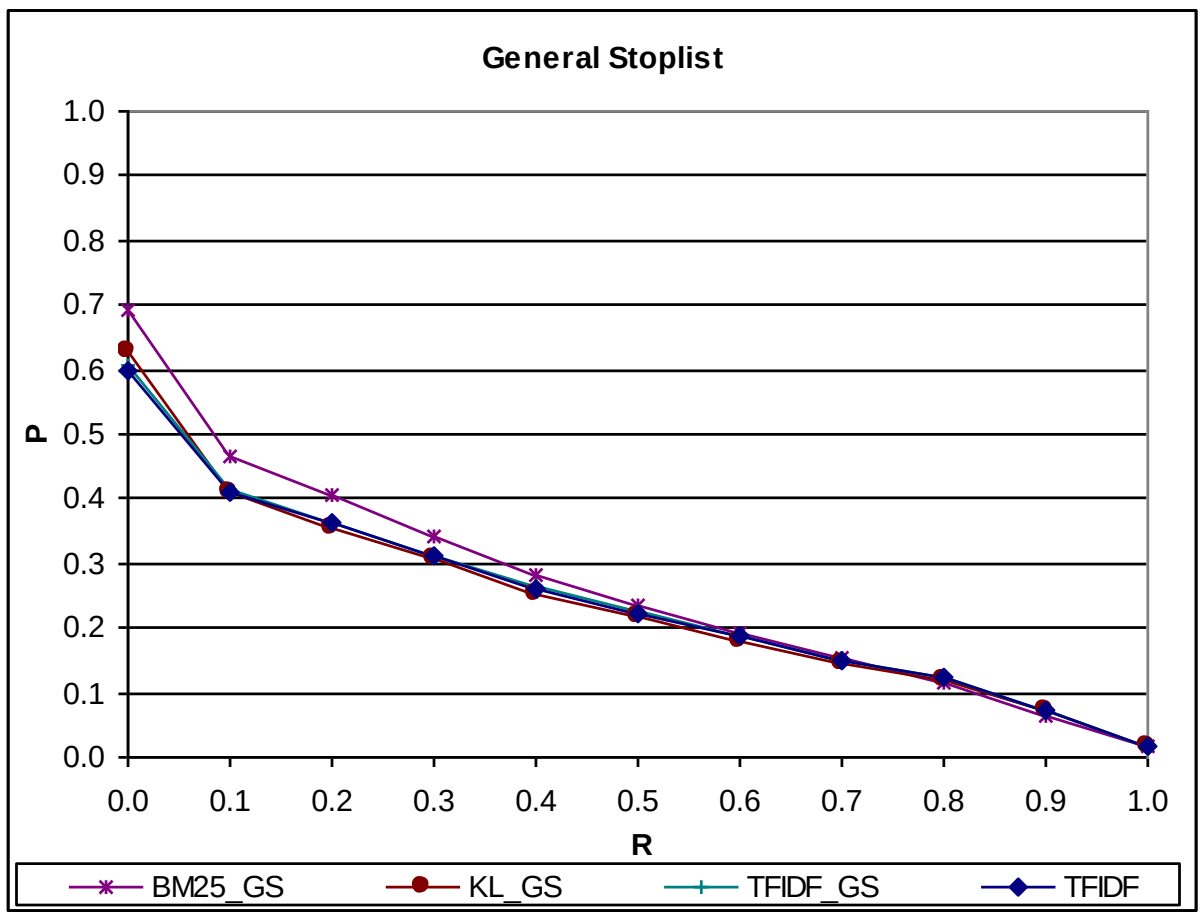

**Figure 2:** Recall-Precision Curves for the General Stoplist

The differences in the mean precision were minimal, but the increase in precision that was made by the list was apparent at low cut-off levels, especially for the BM25. The Wilcoxon test indicates that the differences between the KL_GS and the baseline precision were not statistically significant. There was a minimal improvement for only one query with the KL model. As for the BM25_GS, and TFIDF_GS there was a significant difference and change. After combining the GS with BM25 the results changed drastically going up from 30 queries better than the baseline precision to 47 queries, and the same thing happened with the TFIDF_GS run with 49 queries favoring it over the baseline precision. Both results indicate how sensitive these weights are, especially the BM25, to the use of stopwords, bearing in mind that the term frequency (TF) portion in the TFIDF is calculated using the BM25 term frequency function. Conversely, the KL model had poor performance with the stoplist as the results slightly deteriorated when it was combined with the stoplist.

### *4.2.2. Corpus-Based Stoplist*

The list has improved the precision for the BM25 weight on the lower levels of recall. Comparing these results with the baseline precision, the test statistic for the Friedman test $\chi^2$ is 12.957 and the P-value is .005. This indicates that the differences between these runs and the baseline precision are significant. The differences in the mean precision were minimal.

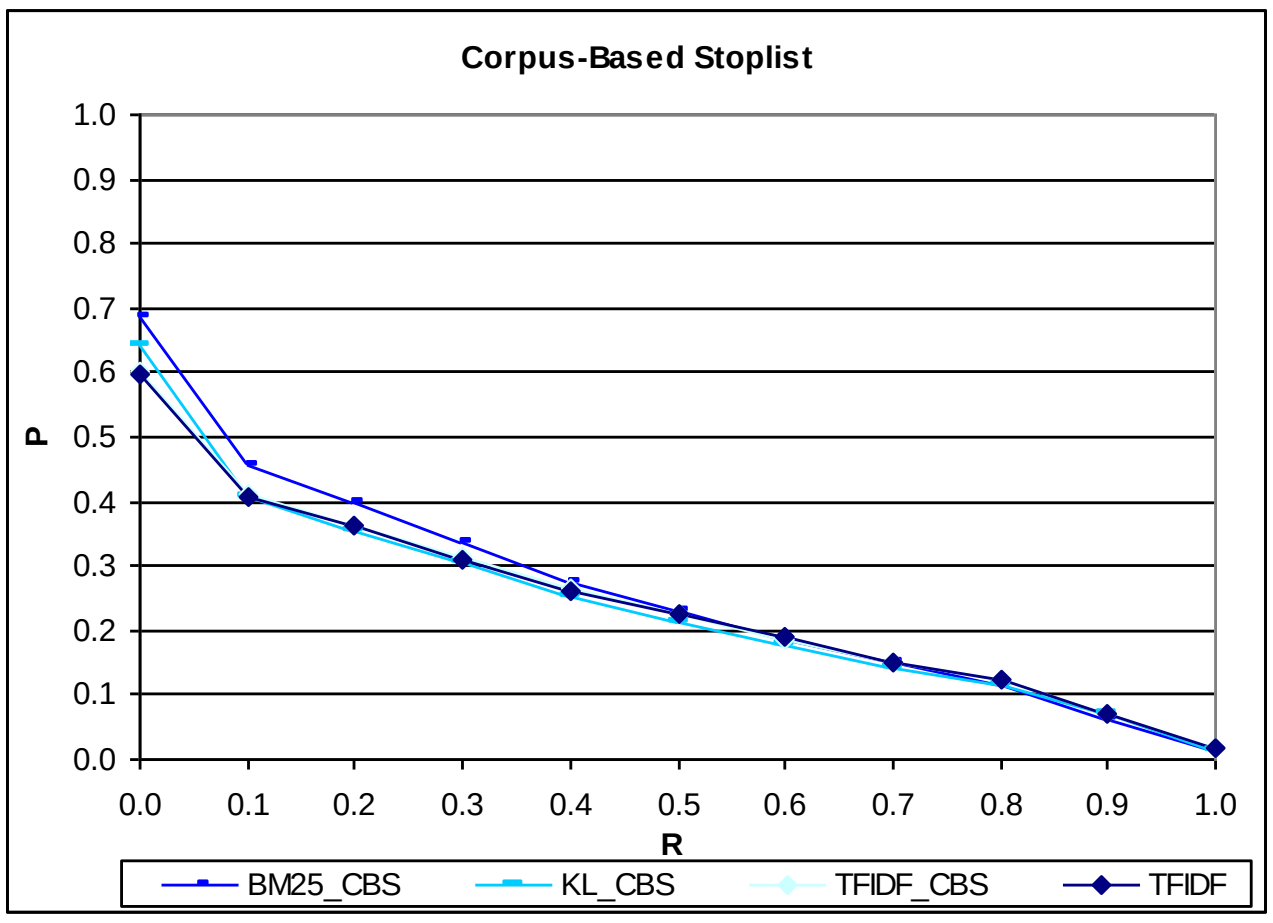

**Figure 3:** Recall-Precision curves for the Corpus-Based Stoplist

The Wilcoxon test indicates that the differences between the BM25_CBS, TFIDF_CBS and the baseline precision were not statistically significant, even though there was an apparent improvement with the BM25. Even though there was a slight improvement with the KL-Model (not more than 2.7 %), the corpus-based stoplist had a negative effect on the overall performance of the model as the results degraded from 31 queries in favor of the KL-Model to 25 when it was combined with the stoplist.

### *4.2.3. Combined Stoplist*

Figure (4) presents the results obtained from using the combined stoplist. In this figure, the curves show that the results were also very close, as for the general stoplist, and that the list has improved the precision for the BM25 weight on the lower levels of recall. Comparing these results with the baseline precision, the test statistic for the Friedman test $\chi^2$ is 13.327 and the P-value is .004. This indicates that the differences between these runs and the baseline precision are significant. The differences in the mean precision were minimal.

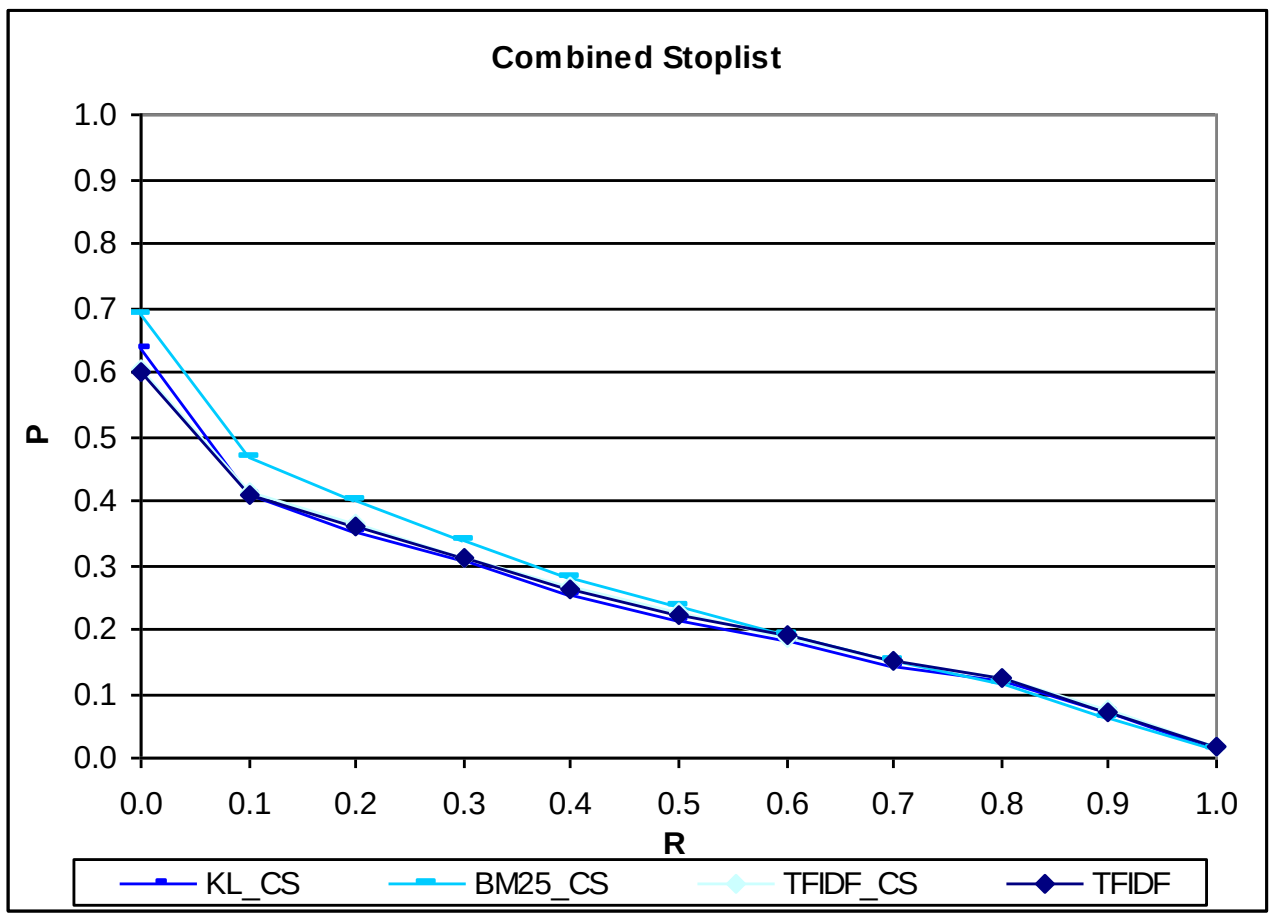

**Figure 4:** Recall-Precision curves for the combined Stoplist

The Wilcoxon test indicates that the differences between the BM25_CS, KL_CS, and the baseline precision were not statistically significant. However there was an improvement for the BM25 weight. Compared to the general stoplist, there were some differences but the results were almost identical to the general stoplist combinations despite the additions to it. Looking at the individual queries the differences in precision were in favor of the general stoplist, showing that the improvement in both was very much due to it. Even though there was an improvement in the KL-Model at low document cut-off levels, the overall performance of the model also had poor performance with the stoplist. The results deteriorated when it was combined with the stoplist. When combined with the TFIDF the combined list had a better overall performance than the baseline precision and the difference was significant.

## 5. Discussion

Using the standard recall and precision measures the above techniques were compared. Six techniques were used separately and combined, generating a total of 24 different indexing approaches.

Without any additional linguistic processing the three schemes, TF*IDF weighting, Okapi best match algorithm, and the Kullback-Leibler Divergence Model, had a good performance with, Arabic which was not surprising considering their previous success with other languages as they depend only on the corpus and query statistics. The differences between the three weighting schemes were very minimal and not statistically significant.

The TF*IDF scheme is the best weighting scheme to be used with the Arabic language when used separately without stemming or stopwords removal. This contradicts some previous research indicating that the BM25 algorithm is better when used with Arabic [17].

One reason for this is that the term frequency portion of the TF*IDF scheme in Lemur is calculated using the term frequency portion in BM25 giving it the advantages of both of schemes. A second is that in Savoy and Rasolofo's experiment the BM25 was combined with a stemmer which particularly boosts the results with Arabic language.

Three stoplists were created for this study, a general stoplist, a corpus-based stoplist, and a combined stolist. The assumption here was that stopwords affect the retrieval process but the extent of this effect was not known. The results showed that this effect varies from one weighting scheme to the other. Combining the stoplists with the TFIDF and KL model did not make a substantial difference, only 0.1%-0.2% increase in the former and decrease in the latter. When the stoplists were combined with the BM25 weight there was noticeable improvement of 7.67% with the corpus-based stoplist, 9.49% with the combined stoplist, and 10.44% with the general stoplist.

**Table 2:** The Wilcoxon Test for All Runs

| Technique | Mean Precision | QP > BP | QP < BP | QP = BP | P-Value |
|---|---|---|---|---|---|
| KL_CBS | 0.2217 | 25 | 49 | 0 | 0.009 |
| KL_CS | 0.2243 | 29 | 45 | 0 | 0.107 |
| BM25 | 0.2169 | 30 | 45 | 0 | 0.093 |
| KL | 0.2264 | 31 | 43 | 1 | 0.287 |
| KL_GS | 0.2248 | 32 | 42 | 0 | 0.229 |
| TFIDF_CBS | 0.226 | 35 | 39 | 1 | 0.668 |
| TFIDF_CS | 0.2285 | 45 | 29 | 1 | 0.037 |
| BM25_CBS | 0.2433 | 46 | 29 | 0 | 0.123 |
| BM25_CS | 0.2475 | 46 | 29 | 0 | 0.056 |
| BM25_GS | 0.2496 | 47 | 28 | 0 | 0.018 |
| TFIDF_GS | 0.2283 | 49 | 25 | 0 | 0.028 |

- QP: Query Precision.
- BP: Baseline Precision (TFIDF).

The results illustrate how sensitive the BM25 weight and KL model are to the use of stopwords. The use of stopwords has positively affected the BM25 weight while associating it with the KL model has affected the results negatively. The corpus-based list was the lower than the General list, which suggests that we should revisit the corpus-based list. Unfortunately there are no clear-cut rules on how to create a list like this and most of the decisions that were taken in creating this list were arbitrary.

Generally the overall performance of the general stoplist was better than the corpus-based stoplist and to some extent better than the combined stoplist. The list can be used as a standard list for Arabic retrieval regardless of the nature of the data used. The list will be added to the Lemur Toolkit making it available for research and further development as there are no publicly available stoplists for Arabic language.

As for the Lemur Toolkit, one of the main objectives of this study was to use it in experimenting with Arabic language retrieval. The addition of the Arabic language to the Lemur Toolkit will benefit the language as it facilitates the retrieval process whatever the approach that is followed. A major advantage of the toolkit is that it is open source software making it easy to add or modify applications. During this study few applications of the toolkit were used but it proved to be very efficient when used to work with the Arabic language in terms of time, the capability to handle the language, and the ease of use.

## 6. Conclusions and Further Research

Experimentation with Arabic language retrieval is still a relatively new area of research; it still requires exploring and more research. In this study several retrieval techniques and their potential in improving retrieval effectiveness were explored. The effects of term weighting, and stopwords on Arabic retrieval were examined and compared using the Lemur Toolkit.

The best match algorithm, BM25, with the combined or general stoplist was the best performing function for retrieval in the Arabic language. The performance of a general stoplist or a combined list was relatively close. The use of any of them is recommended but the general stoplist is certainly preferred if we are dealing with a different corpus.

The Kullback-Leibler Divergence Model had problems performing with stopwords. Further investigation with the model could reveal the extent of this problem especially when dealing with different smoothing algorithms and variant query lengths.

Developing a new weighting algorithm that could use the characteristics of the Best Match algorithm, BM25, and combining it with language specific characteristics may lead to further improvements. For instance, using the syntactical structure of the Arabic sentence in calculating the term weight with the BM25 could improve upon the efficiency of this algorithm.

Lemur comes equipped with several applications; this study has used only a small percentage of these applications. Experimenting with the other applications provided by the toolkit to determine their performance in Arabic is another area that could be explored, for instance the use of feedback in retrieval, summarization…etc.

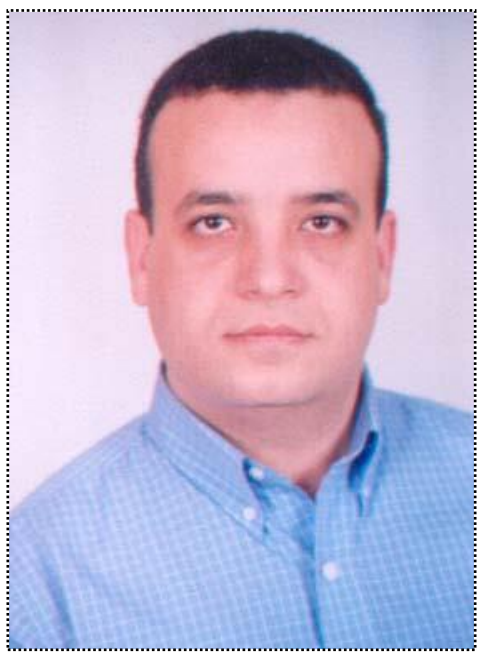

**Dr. Ibrahim Abu El-Khair**
Received MLIS, from the School of Library and Information Science (Currently School of Information Studies) - University of Wisconsin, Milwaukee – USA. 2000

Received Ph.D. in Information Science, School of Information Sciences, University of Pittsburgh – USA. 2003.

Currently working as an assistant professor at the Dept. of Library and Information Science. Faculty of Arts, El-Minia University – Egypt.

## Appendix A. General Stoplist

| | | | | | | | |
|---|---|---|---|---|---|---|---|
| انها | القادم | اياه | أمامهم | أية | إنه | بامكان | بأيهن |
| اثناء | اللاتي | ايضا | أمامهما | أيضا | إنها | بان | بإحدى |
| اجل | اللاحق | اين | أمامهن | أيضاً | إنهم | بانه | بإذا |
| احدا | اللتان | ايها | أمامي | أين | إنهما | باولئك | بإلا |
| احدى | اللتين | آخر | أن | أينما | إنهن | بآخر | بإياك |
| احيانا | اللذان | أبدا | أنا | أيها | إني | بأحد | بإياكم |
| اخرى | اللذين | أثناء | أنت | أيهم | إياك | بأشياء | بإياكما |
| اخيرا | | أجل | أنتم | أيهما | إياكم | بأقل | بإياكن |
| اذ | اللواتي | أحد | أنتما | أيهن | إياكما | بألا | بإياه |
| اذا | المقبل | أحياناً | أنتن | إحدى | إياكن | بأن | بإياها |
| اذن | الممكن | أخرى | أنك | إذ | إياه | بأنا | بإياهم |
| ازاء | المنصر | أخيرا | أنكم | إذا | إياها | بأنك | بإياهما |
| استمرا | م | أخيراً | أنكما | إذاً | إياهم | بأنكم | بإياهن |
| ر | النحو | أزاء | أنكن | إزاء | إياهما | بأنكما | بإياى |
| اصبح | الى | أشياء | أنما | إطلاقاً | إياهن | بأنكن | ببضع |
| اصبحت | اليه | | أننا | إلا | إياى | بأننا | ببضعة |
| اكثر | اليها | أصبح | أنني | إلى | بئس | بأنني | ببعض |
| الا | اليهم | أقل | أنه | إلي | بالأمام | بأنه | ببعضها |
| الان | اما | أكثر | أنها | إليك | بالأمر | بأنها | ببعضهم |
| الآن | امام | ألا | أنهم | إليكم | بالإضاف | بأنهم | بتلك |
| الأمام | امس | ألست | أنهما | إليكما | ة | بأنهما | بحيث |
| الأمر | ان | ألستم | أنهن | إليكن | بالتالي | بأني | بدلا |
| الأن | انا | ألستما | | إلينا | بالتأكيد | بأواخر | بدون |
| الإطلاق | انت | ألستن | أنى | إليه | بالتي | بأولئك | بدوننا |
| البعض | انتم | ألسن | أني | إليها | بالذي | بأولاء | بدونه |
| التى | انك | أليس | أو | إليهم | بالذين | بأولائك | بدونها |
| التي | انكم | أليست | أواخر | إليهما | بالرغم | بأولائكم | بدونهم |
| الجاري | انكن | أليسوا | أولئك | إليهن | بالضبط | بأولائكم | بدونهما |
| الحالي | انما | أم | أولا | إما | بالغير | ا | بدونهن |
| الخ | اننا | أما | أولاء | | بالقول | بأولائك | بذا |
| الذان | انني | أمام | أولائك | إن | باللاتي | ن | بذاك |
| الذى | انه | أمامك | أولائكم | إنا | باللتان | بأى | بذلك |
| الذي | انهم | أمامكم | أولائكما | إنك | باللتين | بأي | بذو |
| الذين | انهما | أمامكما | أولائكن | إنكم | باللذان | بأياً | بذي |
| الرغم | انهن | أمامكن | أى | إنكما | باللذين | بأية | برغم |
| السابق | او | أمامنا | أي | إنما | باللواتي | بأيها | بسبب |
| السواء | اولئك | أمامه | أياً | إننا | بالنسبة | بأيهم | بسوى |
| الغير | اي | أمامها | أيان | إنني | | بأيهما | بشأن |

بشكل
بشيء
بشيئاً
بشيئان
بشيئين
بصورة
بضع
بضعة
بعد
بعدئذ
بعدة
بعدم
بعدها
بعض
بعضا
بعضها
بعضهم
بغض
بغير
بغيرك
بغيركم
بغيركما
بغيركن
بغيرنا
بغيره
بغيرها
بغيرهم
بغيرهما
بغيرهن
بغيري
بك
بكافة
بكل
بكم
بكما
بكن
بكيف
بل
بلا

بلى
بما
بماذا
بمتى
بمزيد
بمزيداً
بمفرده
بمن
بن
بنا
بنحو
بنسبة
به
بهؤلاء
بها
بهاتان
بهاتين
بهذا
بهذان
بهذه
بهذي
بهذين
بهل
بهم
بهما
بهن
بين
بينك
بينكم
بينكما
بينكن
بينما
بيننا
بينه
بينها
بينهم
بينهما
بينهن
بيني
تحته

تقريبا
تقريباً
تقول
تكن
تكون
تكونوا
تلك
تلكم
تلكما
تماما
ثم
ثمة
جدا
جداً
جيدا
حاشا
حالما
حاليا
حالياً
حتما
حتى
حسب
حوالي
حول
حولك
حولكم
حولكن
حولنا
حوله
حولها
حولهم
حولهما
حولهن
حولي
حيث
حيثما
حين
حينئذ
حيناً
حينذاك

حينما
حينه
حينها
خارجاً
خاصا
خاصة
خصو
صاً
خصيصـ
ا
خلا
خلال
خلاله
خلف
خلفك
خلفكم
خلفكما
خلفكن
خلفنا
خلفه
خلفها
خلفهم
خلفهما
خلفهن
خلفي
دائما
دائماً
داخلاً
دون
دونك
دونكم
دونكما
دوننا
دونه
دونها
دونهم
دونهما
دونهن
ذا
ذات

ذاتك
ذاتكما
ذاته
ذاتها
ذاتهم
ذاتهما
ذاتهن
ذاك
ذلك
ذلكم
ذلكما
ذو
ذي
ربما
رغم
رغماً
رقم
سواء
سواءاً
سوف
سوى
شانه
شأنه
شتى
شيء
شيئا
شيئاً
شيئان
شيئين
ضدك
ضدكم
ضدكما
ضدكن
ضدنا
ضده
ضدها
ضدهم
ضدهما
ضدهن
ضدي

ضدين
ضرورة
ضرور
ي
ضروري
اً
ضمن
طالما
طويل
طويلاً
طويلة
طويله
ظل
عام
عامة
عبر
عدا
عدة
عدم
عدمه
عديدة
عسى
على
علي
عليّ
عليك
عليكم
عليكما
عليكن
علينا
عليه
عليها
عليهم
عليهما
عليهن
عما
عن
عنّا
عند
عندئذ

عندك
عندكم
عندكما
عندما
عنده
عندها
عندهم
عندهما
عندهن
عنك
عنكم
عنكما
عنم
عنه
عنها
عنهم
عنهما
عنهن
عني
غير
غيرك
غيركم
غيركما
غيركن
غيرنا
غيره
غيرها
غيرهم
غيرهما
غيرهن
غيري
فاذ
فاذا
فاكثر
فالآن
فالأن
فالتي
فالذي
فالذين
فالغير

فالقول
فاللاتي
فاللتان
فاللتين
فاللذان
فاللذين
فاللواتي
فان
فانك
فاننا
فانه
فانها
فانهم
فاولئك
فأحد
فأقل
فأكثر
فألا
فأما
فأن
فأنا
فأنت
فأنتم
فأنتما
فأنتن
فأنه
فأنهم
فأنى
فأولئك
فأولاء
فأولائك
فأولائكم
فأولائكم
ا
فأولائك
ن
فأى
فأيان
فأين
فأينما

فإذ
فإذا
فإلا
فإلى
فإلي
فإليك
فإليكم
فإليكما
فإليكن
فإلينا
فإليه
فإليها
فإليهم
فإليهما
فإليهن
فإما
فإن
فإنا
فإنك
فإنكم
فإنكما
فإننا
فإنه
فإنها
فإنهم
فإنهما
فإني
فإياك
فإياكم
فإياكما
فإياكن
فإياه
فإياها
فإياهم
فإياهما
فإياهن
فإياى
فبئس
فبالتي
فبالذي
فبالذين
فبالغير
فبالقول
فباللاتي
فباللتان
فباللتين
فباللذان
فباللذين
فباللوات
ي
فبالنسبة
فباولئك
فبألا
فبأولئك
فبتلك
فبحيث
فبذا
فبذاك
فبذلك
فبذي
فبعد
فبعدة
فبك
فبكل
فبكم
فبكما
فبكن
فبما
فبماذا
فبنا
فبنسبة
فبهؤلاء
فبها
فبهاتان
فبهاتين
فبهذا
فبهذان
فبهذه
فبهذين
فبهم
فبهما
فبهن
فبين
فبينك
فبينكم
فبينكما
فبينكن
فبينما
فبيننا
فبينه
فبينها
فبينهم
فبينهما
فبينهن
فبيني
فتحت
فتلك
فثم
فجأة
فجأةً
فحاشا
فحيث
فحيثما
فحين
فحينئذ
فحيناً
فحينذاك
فحينما
فحينه
فحينها
فخلا
فخلال
فدائماً
فذا
فذاك
فذلك
فذو
فذي
فسواء
فسواءاً
فسوف
فسوى
فطالما
فعدا
فعدة
فعدم
فعلا
فعلى
فعليّ
فعليك
فعليكم
فعليكما
فعليكن
فعلينا
فعليه
فعليها
فعليهم
فعليهما
فعليهن
فعن
فعنّا
فعند
فعندئذ
فعندك
فعندكم
فعندكما
فعندما
فعنده
فعندها
فعندهم
فعندهما
فعندهن
فعنك
فعنكم
فعنكما
فعنه
فعنها
فعنهم
فعنهما
فعنهن
فعني
فغير
فغيرك
فغيركم
فغيركما
فغيركن
فغيرنا
فغيره
فغيرها
فغيرهم
فغيرهما
فغيرهن
فغيري
ففوق
ففوقك
ففوقكم
ففوقكما
ففوقكن
ففوقنا
ففوقه
ففوقها
ففوقهم
ففوقهما
ففوقهن
ففي
ففيك
ففيكم
ففيكن
ففيما
ففينا
ففيه
ففيها
ففيهم
ففيهما
ففيهن
فقبل
فقد
فقديماً
فقط
فقلت
فقول
فكالتي
فكالذي
فكالذين
فكالقول
فكاللاتي
فكاللتان
فكاللتين
فكاللذان
فكاللذين
فكاللوات
ي
فكأن
فكأنك
فكأنه
فكأنهم
فكأنهما
فكأنهن
فكثير
فكثيراً
فكذلك
فكل
فكلا
فكلانا
فكلاهما
فكلتا
فكلكم
فكلنا
فكله
فكلها
فكلهم
فكلهن
فكلينا
فكليهما
فكم
فكما
فكي
فكيف
فكيلا
فلا
فلأحد
فلأنه
فلأولئك
فلإحدى
فلإنه
فلبئس
فلتلك
فلدى
فلدي
فلديك
فلديكم
فلديكما
فلدينا
فلديه
فلديها
فلديهم
فلديهما
فلديهن
فلذا
فلذاك
فلذلك
فلذي
فلست
فلستم
فلستما
فلستن
فلسن
فلسوف
فلعدم
فلعل
فلقد
فلك
فلكل
فلكلا
فلكلتا
فلكم
فلكما
فلكن
فلكنك
فلكنه
فلكنهم
فلكنهما
فلكنهن
فلكي
فلكيلا
فلم
فلما
فلماذا
فلمذا
فلن
فلنا
فله
فلهؤلاء
فلها
فلهاتان
فلهاتين
فلهتان
فلهتين
فلهذا
فلهذان
فلهذه
فلهذين
فلهم
فلهما
فلهن
فلو
فلولا
فلولاك
فلولاكم
فلولاكما
فلولاكن
فلولانا
فلولاه
فلولاها
فلولاهم
فلولاهما
فلولاهن
فلولاى
فليس
فليست

فليسوا
فما
فماذا
فماعدا
فمتى
فمثل
فمثلاً
فمثلما
فمدام
فمدة
فمع
فمعاً
فمعك
فمعكم
فمعكما
فمعكن
فمعنا
فمعه
فمعها
فمعهم
فمعهما
فمعهن
فمعي
فمما
فمن
فمنا
فمنذ
فمنك
فمنكم
فمنكما
فمنكن
فمننا
فمنه
فمنها
فمنهم
فمنهما
فمنهن
فمني
فمهما
فنحن
فهؤلاء
فهاتان
فهاتين
فهأنت
فهأنتم
فهأنذا
فهتان
فهتين
فهذا
فهذان
فهذه
فهذي
فهذين
فهل
فهم
فهما
فهن
فهنا
فهناك
فهو
فهي
فوق
فوقك
فوقكم
فوقكما
فوقكن
فوقنا
فوقه
فوقها
فوقهم
فوقهما
فوقهن
فى
في
فيك
فيكم
فيما
فينا
فيه
فيها
فيهم
فيهما
فيهن
فيومئذ
قبل
قبله
قبلها
قد
قديماً
قريبا
كافة
كافياً
كالأن
كالتي
كالذي
كالذين
كالقول
كاللاتي
كاللتان
كاللتين
كاللذان
كاللذين
كاللواتي
كان
كانا
كانت
كانتا
كانوا
كأحد
كأن
كأنك
كأنكم
كأننا
كأنه
كأنها
كأنهم
كأنهما
كأنهن
كأني
كأولاء
كأولائك
كأولائك
م
كأولائك
ما
كأولائك
ن
كأى
كإحدى
كإياك
كإياكم
كإياكما
كإياكن
كإياه
كإياها
كإياهم
كإياهما
كإياهن
كإياى
كبيرا
كتلك
كثير
كثيرا
كثيراً
كذا
كذاك
كذلك
كذو
كسوى
كغير
ككل
كل
كلا
كلانا
كلاهما
كلتا
كلكم
كلما
كلنا
كله
كلها
كلهم
كلهن
كلينا
كليهما
كم
كما
كماذا
كمن
كن
كنا
كنت
كنتم
كنتما
كهؤلاء
كهاتين
كهذا
كهذه
كهذي
كهذين
كونه
كونها
كونوا
كي
كيف
كيلا
لئلا
لا
لابد
لان
لانه
لانها
لانهم
لاولئك
لاي
لآخر
لأحد
لأمام
لأمامك
لأمامكم
لأمامكم
ا
لأمامكن
لأمامنا
لأمامه
لأمامها
لأمامهم
لأمامهم
ا
لأمامهن
لأمامي
لأن
لأنا
لأنك
لأنكم
لأنكما
لأنكن
لأننا
لأنني
لأنه
لأنها
لأنهم
لأنهما
لأني
لأواخر
لأولئك
لأولاء
لأولائك
لأولائكم
لأولائك
ما
لأولائك
ن
لأى
لأي
لأياً
لأية
لأيها
لأيهم
لأيهما
لأيهن
لإحدى
لإياك
لإياكم
لإياكما
لإياكن
لإياه
لإياها
لإياهم
لإياهما
لإياهن
لإياى
لبئس
لبعض
لتلك
لدى
لدي
لديك
لديكم
لديكما
لدينا
لديه
لديها
لديهم
لديهما
لديهن
لذا
لذاك
لذلك
لذو
لذي
لست
لستم
لستما
لستن
لسن
لسوف
لسوى
لعدم
لعل
لغه
لغير
لقد
لك
لكل
لكلا
لكلتا
لكم
لكما
لكن
لكنك
لكننا
لكنه
لكنها
لكنهم
لكنهما
لكنهن
لكني
لكي
لكيلا
للأمام
للأمر
للتي
للذي
للذين
للغاية
لللاتي
لللتان
لللتين
لللذان
لللذين
لللواتي
للمزيد
لم
لما
لماذا
لمدة
لمذا
لمزيد
لمزيداً

لمن
لن
لنا
له
لهؤلاء
لها
لهاتان
لهاتين
لهتان
لهتين
لهذا
لهذان
لهذه
لهذي
لهذين
لهم
لهما
لهن
لو
لولا
لولاك
لولاكم
لولاكما
لولاكن
لولانا
لولاه
لولاها
لولاهم
لولاهما
لولاهن
لولاى
لي
ليس
ليست
ليسوا
ليكون
مؤكداً
ما
مادام
ماذا
مازال
مازالت
ماعدا
ماهو
متى
مثل
مثلا
مثلاً
مثلما
مثله
مثلها
مثلهم
مدة
مدى
مرة
مزيد
مزيداً
مطلقاً
مع
معا
معاً
معظم
معك
معكم
معكما
معكن
معنا
معه
معها
معهم
معهما
معهن
معي
مم
مما
ممكن
ممكناً
ممن
من
منا
منذ
منك
منكم
منكما
منكن
مننا
منه
منها
منهم
منهما
منهن
مني
مهما
نحن
نظرا
نعم
هؤلاء
هاتان
هاتين
هاذين
هامة
هأنت
هأنتم
هأنذا
هذا
هذان
هذه
هذي
هذين
هكذا
هل
هم
هما
هن
هنا
هناك
هنالك
هو
هي
وراء
وراءه
ورائك
ورائكم
ورائكما
ورائكن
ورائهم
ورائهما
ورائهن
يا
يبدو
يكن
يكون
يكونوا
يلي
يمكن
يمكنه
يومئذ

# Appendix B. Corpus-Based Stoplist

ابر
ابو
اجتماع
اجل
احد
اخرى
اذا
ارا
اربعة
اط
اطار
اطلاق
اعادة
اعلن
اغو
اف
افب
اكثر
اكد
الا
الاتفاق
الاثنين
الاحد
الاخيرة
الاراضي
الاربعاء
الاسبوع
الان
الانباء
الاول
الاولى
التعاون
التى
التي
الثانية
الثلاثاء
الجمعة
الجيش
الحالي
الحدود
الحزب
الحكم
الخميس
الدولة
الدولية
الذى
الذي
الذين
السابق
الساعة
السبت
السلطات
السلطة
الشرطة
العاصمة
العسكرية
العمل
الف
القدم
اللجنة
الماضي
المباراة
المتحدث
المجلس
المجموعة
المرحلة
المصدر
المقبل
المقرر
الملك
النار
النهائي
الوزراء
الوضع
الوقت
الى
اليوم
اما
امام
امس
ان
انه
انها
او
اوك
اول
اي
ايار
ايام
ايضا
أن
باسم
بان
برس
بسبب
بشكل
بطولة
بعد
بعض
بن
به
بها
بيان
بين
تشرين
تم
ثلاثة
ثم
جمت
جميع
جنوب
جهة
حال
حاليا
حتى
حزيران
حول
حيث
حين
خلال
دورة
دولار
دولة
دون
ديس
ديسك
ذكرت
ذلك
رئيس
زيارة
سبت
سنوات
شخصا
شرق
صباح
صحيفة
صفر
ضمن
عام
عاما
عبد
عبر
عدة
عدد
عدم
عشر
عشرة
على
علي
عليه
عليها
عمان
عن
عند
عندما
غدا
غير
فان
فبر
فرانس
فسب
فى
في
فيه
فيها
قال
قبر
قبرص
قبل
قد
قدم
قرار
قمة
قوات
كان
كانت
كانون
كأس
كبير
كرة
كل
كما
لا
لدى
لقاء
لكرة
لكن
للامم
لم
لن
له
لها
لوكالة
مؤتمر
ما
مار
ماي
مايو
مباراة
مجموعة
مدينة
مساء
مصادر
مصدر
مع
مليون
من
منذ
منها
موا
موسع
نحو
نفسه
نقطة
نهاية
نوف
هان
هذا
هذه
هناك
هو
هي
وح
وزارة
وزير
وكالة
يتبع
يجب
يذكر
يكون
يمكن
ينا
يول
يون
يونيو